# Comparing Time and Frequency Domain for Audio Event Recognition Using Deep Learning

Lars Hertel*, Huy Phan*[†] and Alfred Mertins*
*Institute for Signal Processing, University of Luebeck, Germany
[†]Graduate School for Computing in Medicine and Life Sciences, University of Luebeck, Germany
Email: {hertel, phan, mertins}@isip.uni-luebeck.de

*Abstract*—Recognizing acoustic events is an intricate problem for a machine and an emerging field of research. Deep neural networks achieve convincing results and are currently the state-of-the-art approach for many tasks. One advantage is their implicit feature learning, opposite to an explicit feature extraction of the input signal. In this work, we analyzed whether more discriminative features can be learned from either the time-domain or the frequency-domain representation of the audio signal. For this purpose, we trained multiple deep networks with different architectures on the Freiburg-106 and ESC-10 datasets. Our results show that feature learning from the frequency domain is superior to the time domain. Moreover, additionally using convolution and pooling layers, to explore local structures of the audio signal, significantly improves the recognition performance and achieves state-of-the-art results.

## I. INTRODUCTION

Recognizing acoustic events in natural environments, like gunshots or police sirens, is an intricate task for a machine. The effortlessness of the human ear and brain deceives the complex underlying process. However, having a machine that understands its environment, e.g. through acoustic events, is important for many applications such as security surveillance and ambient assisted living, especially in an aging population. This is one reason why machine hearing is becoming a more and more emerging field of research [1].

So far, most of the audio event recognition systems have used hand-crafted features, extracted from the frequency domain of the audio signal. They are mainly borrowed from the field of speech recognition, such as mel-scale filter banks [2], log-frequency filter banks [3] and time-frequency filters [4]. However, with the rapid advance in computing power, feature learning is becoming more common [5]–[7].

In this work, we use deep neural networks in general and convolutional networks in particular for combined feature learning and classification. They have been succesfully applied to many different pattern recognition tasks [8]–[11], including audio event recognition [5], [6], [12], [13]. A schematic representation of a one-dimensional convolutional neural network is shown in Figure 1. The given network comprises five different layers, i.e. input, convolution, pooling, fully connected, and output layers. Given an input signal in the input layer, multiple filters are learned and convolved with the input signal in the convolution layer, resulting in various convolved signals. Multiple values of those signals are then pooled together in the pooling layer. This introduces an invariance to small translations of the input signal. Both convolution and pooling layers are usually applied multiple times. Afterwards, the extracted features are weighted and combined in the fully-connected layer and output in the output layer. There typically exists one output neuron for each audio event category in the output layer.

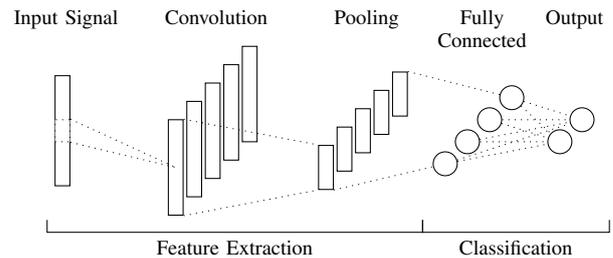

Fig. 1. Schematic diagram of a one-dimensional convolutional neural network for audio event recognition. The network comprises five different layers. Both feature extraction and classification are learned during training.

The motivational question we want to answer in this paper is whether more discriminative features can be learned from the time-domain or the frequency-domain representation of the audio input signal. For this purpose, we train various deep neural networks with different architectures on multiple datasets both in time and frequency domain and compare their achieved recognition results.

## II. DATASETS

To train and evaluate our deep networks, we used two different datasets, namely Freiburg-106 and ESC-10. Both datasets contain short sound clips of isolated environmental audio events. Note that the audio events are not overlapping. There is only a single event present in each sound file. In the following, we will briefly introduce both datasets. An overview of some statistics of the two datasets before and after preprocessing is given in Table I.

### A. Freiburg-106

The Freiburg-106 [14] dataset contains 1,479 audio-based human activities of 22 categories with a total duration of 48 min. It was collected using a consumer-level dynamic cardioid microphone. The audio signals were preamplified and sampled at 44 100 Hz. Several sources of stationary ambient noise were present. The average duration of a recording is 1.9 s. We split the dataset into a training and test set of equal size, i.e. every other recording was used for testing[1].

TABLE I
STATISTICS OF THE USED DATASETS.

| Dataset | Classes | Duration | | Samples | |
|---|---|---|---|---|---|
| | | Total (min) | Average (s) | Training | Test |
| Freiburg-106 | 22 | 48 | 1.90 | 763 | 755 |
| Audio Frames | | | | 129,320 | 133,043 |
| ESC-10 | 10 | 33 | 5.00 | 320 | 80 |
| Audio Frames | | | | 142,101 | 35,606 |

*B. ESC-10*

The ESC-10 [15] dataset contains 400 environmental recordings of 10 classes with a total duration of 33 min. The recordings are uniformly distributed, i.e. 40 recordings for each class. They were searched, downloaded, verified and annotated by Piczak [15] from the publicly available freesound[2] database. Afterwards, short sound clips of 5 s were extracted, resampled to 44 100 Hz and stored with a bitrate of 192 kbit/s using Ogg Vorbis compression. The dataset is split into five parts for a five-fold cross validation. The average human classification accuracy is 95.7 % [15].

*C. Preprocessing*

Before being able to train our networks, we had to preprocess all audio files to a unified format. First, we converted all stereo audio files to mono by averaging the two channels. This was necessary, since some audio files were only mono recordings. Secondly, to reduce the amount of data while maintaining most of the important frequencies, we resampled the audio files to a sampling frequency of 16 000 Hz. Thirdly, we changed the audio bit depth from their original formats to 32 bit floating points and scaled the amplitudes to the range of $[-1, 1]$. Fourthly, we applied a rectangular sliding window to each audio file with a window size of 150 ms and a step size of 5 ms. Thus, audio frames with a fixed size of 2,400 samples were extracted. The window size was determined via a validation set. Applying a sliding window was necessary since deep neural networks insist on a fixed input size. When we trained our networks in the frequency domain, we used a Hamming window instead of a rectangular one, calculated the Fourier transform and concatenated the first half of both the symmetric magnitude and phase of the Fourier transform. Thereby, the network inputs in both time and frequency domain were equally sized with a fixed length of 2,400 samples. Note that by calculating the Fourier transform, we do not lose any information, since the original audio signal can be recovered with the inverse Fourier transform.

[1]This is based on unofficial communication with Stork et al. [14]
[2]http://www.freesound.org

TABLE II
ARCHITECTURE OF OUR IMPLEMENTED DEEP NETWORKS.

| No. | Layer | Dimension | Probability | Parameters |
|---|---|---|---|---|
| 0 | Input | 2,400 | - | - |
| 1 | Dropout | 2,400 | 0.2 | - |
| 2 | Fully Connected | 384 | - | 921,984 |
| 3 | Dropout | 384 | 0.5 | - |
| 4 | Fully Connected | 384 | - | 147,840 |
| 5 | Dropout | 384 | 0.5 | - |
| 6 | Fully Connected | 384 | - | 147,840 |
| 7 | Dropout | 384 | 0.5 | - |
| 8 | Fully Connected | 384 | - | 147,840 |
| 9 | Dropout | 384 | 0.5 | - |
| 10 | Fully Connected | 384 | - | 147,840 |
| 11 | Dropout | 384 | 0.5 | - |
| 12 | Fully Connected | $x$ | - | $x$ |
| 13 | Softmax | $x$ | - | - |

III. METHODS

We then trained both a standard deep neural network and a convolutional network on Freiburg-106 and ESC-10 in both time and frequency domain of the audio events. Consequently, we trained eight deep networks in total.

*A. Deep Network*

The architecture for the standard deep network is shown in Table II. The network comprises 14 layers with more than 1.5 million trainable weights. The input layer 0 expects a signal with 2,400 values, corresponding to a single audio frame. The number of neurons for the output layer 15 depends on the number of classes, i.e. 22 for Freiburg-106 and 10 for ESC-10. To obtain a probability distribution of $n$ output values $\boldsymbol{x}$, we employed the softmax function in layer 15:

$$\text{softmax}(\boldsymbol{x})_i = \frac{\exp(x_i)}{\sum_{j=1}^{n} \exp(x_j)} \text{ for } i = 1, \ldots, n. \quad (1)$$

Between input and output layer we used five fully connected hidden layers. We chose the rectified linear unit (relu) as a nonlinear activation function of an output value $x$:

$$\text{relu}(x) = \max(0, x). \quad (2)$$

Glorot et al. [16] showed its advantages over the sigmoid and hyperbolic tangent as nonlinear activation functions. To prevent the network from overfitting, we regularized it by using dropout [17] after each layer. The probability to randomly drop a unit in the network is 20 % for the input layer and 50 % for all the hidden layers. Moreover, we used a maximum norm constraint $\|w\|_2 < 1$ for any weight $w$ in the network, as suggested by Hinton [18]. This form of regularization bounds the value of the weights while not driving them to be near zero, as e.g. in weight decay.

*B. Convolutional Network*

The architecture for our convolutional network is shown in Table III. The network comprises 16 layers with nearly

TABLE III
ARCHITECTURE OF OUR IMPLEMENTED CONVOLUTIONAL NETWORKS.

| No. | Layer | Dimension | | Size | Stride | Parameters |
|---|---|---|---|---|---|---|
| | | Rows | Columns | | | |
| 0 | Input | 1 | 2,400 | - | - | - |
| 1 | Dropout | 1 | 2,400 | - | - | - |
| 2 | Convolution | 48 | 2,392 | 9 | 1 | 480 |
| 3 | Pooling | 48 | 598 | 4 | 4 | - |
| 4 | Convolution | 96 | 590 | 9 | 1 | 41,568 |
| 5 | Pooling | 96 | 147 | 4 | 4 | - |
| 6 | Convolution | 192 | 139 | 9 | 1 | 166,080 |
| 7 | Pooling | 192 | 34 | 4 | 4 | - |
| 8 | Convolution | 384 | 26 | 9 | 1 | 663,936 |
| 9 | Pooling | 384 | 6 | - | - | - |
| 10 | Fully Connected | 1 | 384 | - | - | 885,120 |
| 11 | Dropout | 1 | 384 | - | - | - |
| 12 | Fully Connected | 1 | 384 | - | - | 147,840 |
| 13 | Dropout | 1 | 384 | - | - | - |
| 14 | Fully Connected | 1 | $x$ | - | - | $x$ |
| 15 | Softmax | 1 | $x$ | - | - | - |

2 million trainable parameters. The input and output layer are identical to the standard deep network. However, in between we additionally have convolution and pooling layers. In the convolution layer, the input signal is convolved with multiple learned filters of a fixed size with a fixed stride using shared weights. We used a filter size of 9, analogous to 3×3 filters that are often used in computer vision. The number of learned kernels are 48, 96, 192, and 384, respectively. Note that after the first convolution our one-dimensional input signal does not become a two-dimensional image, but multiple one-dimensional signals (c.f. Figure 1). Hence, we only applied one-dimensional convolutions. The pooling layer then reduces the size of the signal while trying to maintain the contained information and introducing an invariance to small translations of the input signal. The pooling size and stride was set to 4, analogous to 2×2 pooling that is again often used in computer vision. We used maximum pooling for all pooling layers. As a nonlinear activation function, we again settled for the rectified linear unit, just as in standard deep networks. Afterwards, the extracted features from the input signal were combined using three fully connected layers. To regularize our network, we again used dropout layers. This time, however, dropout was only used after the input layer with a probability of 20 % and after each fully connected layer with a probability of 50 %.

We used the Python library *Theano* [19], [20] and the NVIDIA *CUDA Deep Neural Network*[3] (cuDNN v3) library to train our deep networks. The library allowed us to employ the GPU[4] of our computer for faster training. This resulted in a speedup of approximately ten, compared to training on the CPU[5].

The standard deep neural networks were trained for 100 epochs. An epoch means a complete training cycle over all audio frames of the training set. One single epoch took nearly 30 s. We started with a fixed learning rate of 0.05 and decreased it by a factor of two after 20 epochs. Furthermore, we selected a batch size of 256 frames and a momentum of 0.9. In constrast, the convolutional networks, were trained for 20 epochs. A single epoch took nearly 11 min. We again started with a fixed learning rate of 0.05 and decreased it by a factor of two after five epochs. Batch size and momentum remained the same as for standard deep networks.

To predict the class label of an entire audio file $\boldsymbol{X}$ of our test set, we first predicted each of the $n$ audio frames individually. Due to the softmax output layer of our network we obtained a probability distribution among the $m$ class labels. Afterwards, we performed a probability voting by adding the predicted probabilities for each frame together and taking the class label with the maximum probability:

$$\text{vote}(\boldsymbol{X}) = \underset{j=1,\ldots,m}{\arg\max}\left(\sum_{i=1}^{n} x_{ij}\right). \quad (3)$$

To evaluate our predicted class labels, we used the f-score metric:

$$\text{f-score} = 2 \cdot \frac{\text{precision} \cdot \text{recall}}{\text{precision} + \text{recall}}, \quad (4)$$

which considers both precision and recall values and can be interpreted as the weighted average of the precision and recall.

IV. RESULTS

Our results are given in Figure 2, Table IV and Table V. For comparison, the state-of-the-art results are 98.3 % [21] for Freiburg-106 and approximately 80 %[6] [15] for ESC-10. The human accuracy for ESC-10 is 95.7 % [15].

Figure 2 displays the average f-score in percent for the standard deep neural networks on the validation test set. The solid lines represent training in the frequency domain and the dashed lines represent training in the time domain for both Freiburg-106 and ESC-10, respectively. Note that the shown f-score was calculated and averaged for a single audio frame, not an entire audio file. Thus, no voting had been performed yet. Clearly, audio events in Freiburg-106 are easier to recognize than in ESC-10. Moreover, for both datasets, networks trained in the frequency domains achieved a higher f-score than networks trained in the time domain.

More detailed results for Freiburg-106 are given in Table IV. It shows the f-score for each individual audio event category and the average f-score value, obtained with probability voting. Standard deep neural networks reach an average f-score of 75.9 % in the time domain and 97.6 % in the frequency domain. Convolutional networks, however, reach an overall accuracy of 91.0 % in time domain and 98.3 % in the frequency domain. The improvement in the time domain is therefore 15.1 % and 0.7 % in the frequency domain. The

---

[3]https://developer.nvidia.com/cudnn
[4]GeForce GT 640 with 2 GB of memory
[5]Intel Core i7-3770K with eight cores
[6]The recognition results are only given in form of a boxplot.

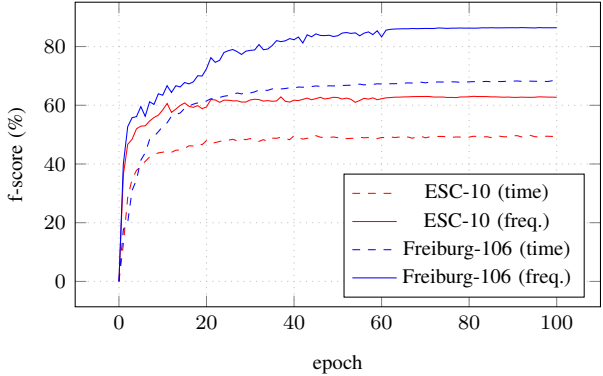

Fig. 2. Comparing the validation f-score of multiple standard deep neural networks on two datasets. The networks were trained for 100 epochs. The solid lines represent training in the frequency domain and the dashed lines represent training in the time domain, respectively.

TABLE IV
RECOGNITION RESULTS FOR THE FREIBURG DATASET (F-SCORE IN %).

| No. | Class | Deep Network | | Convolutional Network | |
|---|---|---|---|---|---|
| | | Time | Frequency | Time | Frequency |
| 0 | Background | 32.3 | 78.0 | 45.8 | 75.0 |
| 1 | Bag | 77.5 | 98.8 | 95.0 | 100.0 |
| 2 | Blender | 95.1 | 100.0 | 100.0 | 100.0 |
| 3 | Cornflakes Bowl | 75.9 | 100.0 | 72.2 | 100.0 |
| 4 | Cornflakes Eating | 86.4 | 100.0 | 95.2 | 100.0 |
| 5 | Cup | 14.4 | 95.7 | 90.9 | 100.0 |
| 6 | Dish Washer | 93.7 | 97.8 | 100.0 | 100.0 |
| 7 | Electric Razor | 96.3 | 97.6 | 100.0 | 100.0 |
| 8 | Flatware Sorting | 46.7 | 97.6 | 50.0 | 100.0 |
| 9 | Food Processor | 86.7 | 100.0 | 94.1 | 100.0 |
| 10 | Hair Dryer | 90.4 | 100.0 | 100.0 | 100.0 |
| 11 | Microwave | 98.9 | 100.0 | 100.0 | 100.0 |
| 12 | Microwave Bell | 95.7 | 100.0 | 91.6 | 100.0 |
| 13 | Microwave Door | 33.3 | 97.7 | 65.1 | 91.3 |
| 14 | Plates Sorting | 59.1 | 98.5 | 86.6 | 100.0 |
| 15 | Stirring Cup | 89.7 | 98.3 | 100.0 | 100.0 |
| 16 | Toilet Flush | 70.0 | 95.8 | 88.7 | 96.8 |
| 17 | Toothbrush | 64.6 | 96.3 | 85.7 | 100.0 |
| 18 | Vacuum Cleaner | 90.9 | 100.0 | 100.0 | 100.0 |
| 19 | Washing Machine | 92.4 | 98.5 | 97.0 | 100.0 |
| 20 | Water Boiler | 94.0 | 100.0 | 96.9 | 100.0 |
| 21 | Water Tap | 85.2 | 96.6 | 96.3 | 100.0 |
| | Average | 75.9 | 97.6 | 91.0 | 98.3 |

*background* class was most difficult to recognize by the networks, while nearly all audio events of the *Microwave* category were correctly recognized by all the different networks.

As for the recognition results for the ESC-10 dataset in Table V, standard deep neural networks reach an average f-score of 70.3 % with training in the time domain and 77.1 % in the frequency domain. Convolutional networks improve these results by 13.4 % to 83.7 % in the time domain and by 12.8 % to 89.9 % in the frequency domain, respectively. Nearly all audio events of the *dog bark* class were correctly recognized by all the different networks, while recognizing a *chainsaw* was most difficult in the frequency domain and *sea waves* most difficult in the time domain, respectively.

TABLE V
RECOGNITION RESULTS FOR THE ESC-10 DATASET (F-SCORE IN %).

| No. | Class | Deep Network | | Convolutional Network | |
|---|---|---|---|---|---|
| | | Time | Frequency | Time | Frequency |
| 0 | Baby Cry | 62.5 | 76.2 | 93.3 | 100.0 |
| 1 | Chainsaw | 80.0 | 71.4 | 75.0 | 71.4 |
| 2 | Clock Tick | 66.6 | 80.0 | 84.2 | 80.0 |
| 3 | Dog Bark | 87.5 | 100.0 | 100.0 | 100.0 |
| 4 | Fire Crackling | 54.5 | 40.0 | 85.7 | 80.0 |
| 5 | Helicopter | 94.1 | 88.9 | 94.1 | 100.0 |
| 6 | Person Sneeze | 50.0 | 66.7 | 71.4 | 80.0 |
| 7 | Rain | 61.5 | 85.7 | 66.7 | 94.1 |
| 8 | Rooster | 76.9 | 85.7 | 100.0 | 100.0 |
| 9 | Sea Waves | 69.6 | 76.2 | 66.7 | 93.3 |
| | Average | 70.3 | 77.1 | 83.7 | 89.9 |

## V. DISCUSSION

Deep convolutional networks are the state-of-the-art approach for many pattern recognition tasks, including audio event recognition. One reason is the implicit feature learning instead of an explicit feature extraction of the input signal. In this work, we analyzed whether more suitable features can be learned from either the time domain or the frequency domain.

Our results show that learning from the frequency domain is consistently superior to learning from the time domain on both datasets Freiburg-106 and ESC-10. Our trained deep neural networks achieved state-of-the-art results. Accordingly, more discriminative features could be learned in the frequency domain.

Moreover, additionally adding convolution and pooling layers to the deep neural network could most of the time significantly improve the achieved f-score. One exception is for learning in the frequency domain on Freiburg-106, where a standard deep network alone already reached comparable state-of-the-art results. Thus, exploring local structures of the input signal both in time and frequency domain seems reasonable.

When training deep networks for audio event recognition, we experienced heavy overfitting of the networks, especially when trained in the time domain. Therefore, we had to intensively regularize the network by employing dropout in each layer. Additionally, we constrained the norm of each weight, as suggested by Hinton [18]. Its main advantage over other regularization methods, like weight decay for example, is that it does not drive the weights to be near zero. This partly prevented the networks from overfitting. However, overfitting to a small extent was still noticeable.

We experienced that some classes were extraordinarily difficult to recognize, e.g. the *background* class in Freiburg-106. When listening to the audio files of those classes, we noticed that most of the time either a long silence was present in these files or no generic pattern was recognizable. A careful filtering of these files could improve the overall recognition accuracy and should be considered.

As already indicated, we determined the window size of 150 ms by employing a validation set that was split from the training data. We noticed that a too small window size, i.e. below 50 ms, could not grasp the important information contained in the audio signal. A too large window, however, required many parameters in the first fully connected layer of our standard deep neural networks, thus resulting in a long training time. A window size of 150 ms was a reasonable compromise between accuracy and training time.

When training our networks in the frequency domain, we used both the magnitude and phase information of the Fourier transform. The main reason for this was to maintain the same number of input samples that were used for the time domain signal. Consequently, we were able to use the same network architecture in both time and frequency domain. Not too surprisingly, when we removed the phase information, the recognition results of our networks remained the same. In contrast, when training with the phase information only, the networks kept guessing randomly.

Instead of using a rectified linear unit (2) as a nonlinear activation function, we also tested maxout networks [22] with a pooling size of 5. We did not notice any differences in our obtained recognition results, however. Since maxout networks are computationally more expensive than rectified linear units, we settled for the latter.

Furthermore, besides using probability voting (3), we also tried majority voting. For this purpose, we predicted the individual class label for each audio frame and assigned the most frequently predicted class label to the audio file. Our results, however, indicated that probability voting is more appropriate for audio event recognition than majority voting.

## VI. Conclusions

Deep learning is suitable for audio event recognition in both the time domain and the frequency domain of the audio signal. However, more discriminative features are learned by the network in the frequency domain, achieving superior results. Exploring the local structure of audio signals by employing convolution and pooling layers additionally improves the recognition performance of the networks, which then achieve state-of-the-art results. Further research will focus on visualizing and understanding what our deep networks have learned both from the time-domain and frequency-domain representation.